\pdfoutput=1
\documentclass[11pt]{article}

\usepackage{algorithm}  
\usepackage{algpseudocode} 
\usepackage{amsfonts}
\usepackage{acl} 
\usepackage{times}
\usepackage{graphicx}
\usepackage{amsmath}
\usepackage[utf8]{inputenc}
\usepackage[T1]{fontenc}
\usepackage{microtype}
\usepackage{inconsolata}
\usepackage[normalem]{ulem}
\usepackage{hyperref}
\usepackage{multirow}
\usepackage{booktabs}
\usepackage{hyperref}  
\usepackage[utf8]{inputenc} 

\title{Unveiling Hidden Collaboration  within Mixture-of-Experts in Large Language Models} 

\author{  
  Yuanbo Tang\textsuperscript{1},   
  Yan Tang\textsuperscript{2},  
  Naifan Zhang\textsuperscript{1},  
  Meixuan Chen\textsuperscript{1},   
  Yang Li\textsuperscript{1,*} \\[5pt]  
  \small  
  \textsuperscript{1} Tsinghua University International Campus Phase I, Nanshan District, Shenzhen \\
  \small
  \textsuperscript{2} College of Software, Northeastern University, Shenyang, Liaoning Province \\
  \small
  \textsuperscript{*} Corresponding author: \href{mailto:tori2011@gmail.com}{tori2011@gmail.com}  
} 

\begin{document}
\maketitle

\begin{abstract}

Mixture-of-Experts based large language models (MoE LLMs) have shown significant promise in multitask adaptability by dynamically routing inputs to specialized experts. Despite their success, the collaborative mechanisms among experts are still not well understood, limiting both the interpretability and optimization of these models. In this paper, we focus on two critical issues: (1) identifying expert collaboration patterns, and (2) optimizing MoE LLMs through expert pruning. To address the first issue, we propose a hierarchical sparse dictionary learning (HSDL) method that uncovers the collaboration patterns among experts. For the second issue, we introduce the Contribution-Aware Expert Pruning (CAEP) algorithm, which effectively prunes low-contribution experts. Our extensive experiments demonstrate that expert collaboration patterns are closely linked to specific input types and exhibit semantic significance across various tasks. Moreover, pruning experiments show  that our
approach improves overall performance by 2.5\% on average, outperforming existing methods. These findings offer valuable insights into enhancing the efficiency and interpretability of MoE LLMs, offering a clearer understanding of expert interactions and improving model optimization.

\end{abstract}

\section{Introduction}
In recent years, the MoE LLMs have gained significant attention as a computationally efficient framework, demonstrating exceptional representational power for large-scale machine learning tasks \cite{jiang_mixtral_2024,fedus_switch_2022}. By leveraging a dynamic routing mechanism, MoE enables the collaborative operation of specialized "Experts", each designed to process complex input data. Compared to traditional architectures, MoE LLMs offer more flexible and adaptive knowledge representations while reducing computational costs, making them well-suited for resource-intensive situations \cite{cai_survey_2024}. 

\setlength{\belowcaptionskip}{-0.3cm} 
\begin{figure}[t]
    \centering
    \includegraphics[width=\columnwidth]{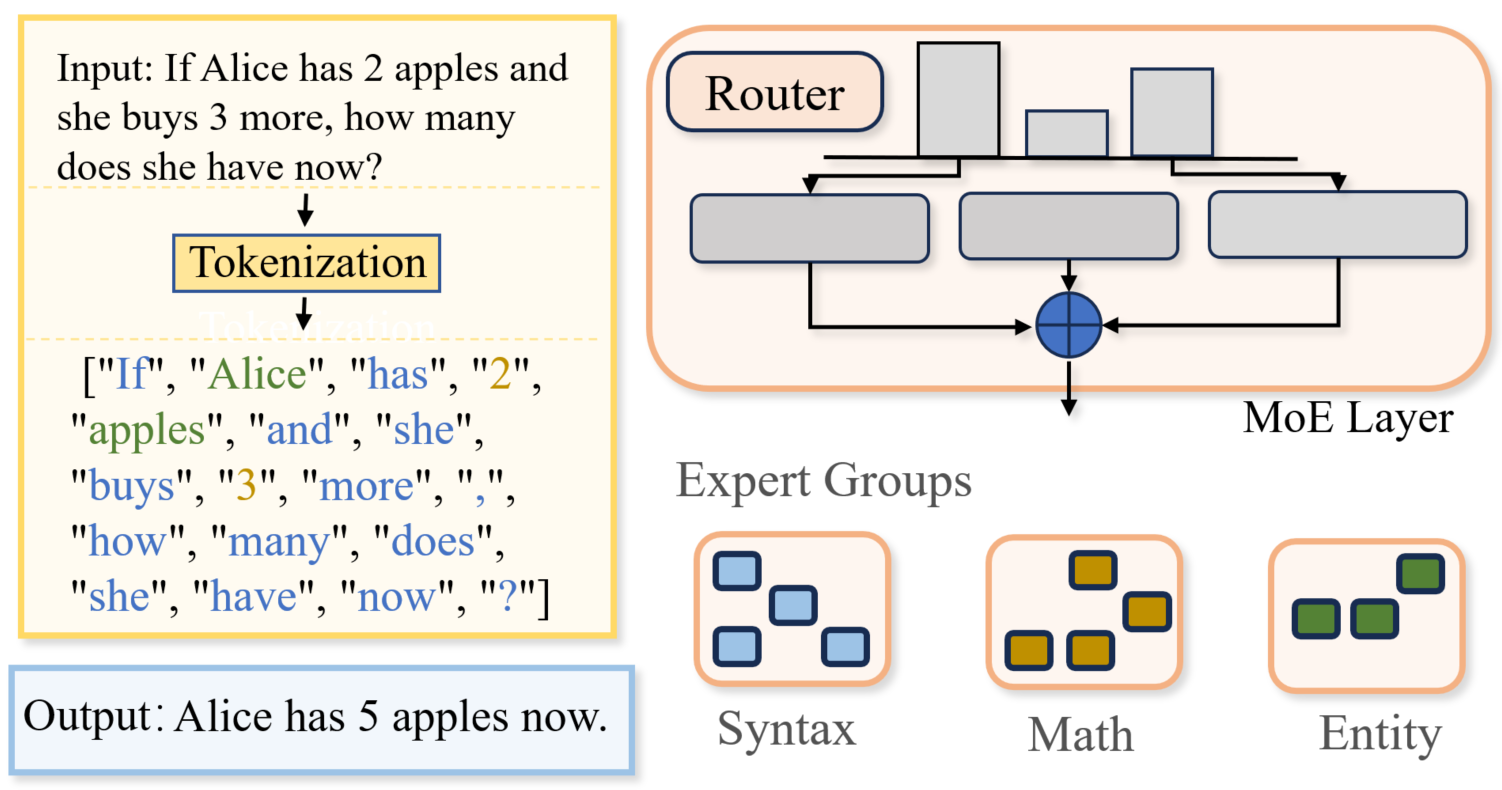}
    \caption{In MoE LLMs, a group of experts often collaborate to analyze a certain type of tokens, and they are not necessarily in the same layer.}
    \label{fig:expert_collaboration}
\end{figure}

Existing research on understanding the working mechanism of MoE LLMs has largely focused on analyzing the behavior of the router, which governs expert selection \cite{lo_closer_2024}. For instance, some studies highlight the influence of output norms on expert selection \cite{lo_closer_2024}, while others reveal that token IDs play a significant role in routing decisions \cite{jiang_mixtral_2024, xue_openmoe_2024, dai_deepseekmoe_2024}. These efforts have provided valuable insights into how MoE LLMs allocate tasks to specialized experts, enhancing multitask adaptability.

\begin{figure}[t]
    \centering
    \includegraphics[width=0.8\columnwidth]{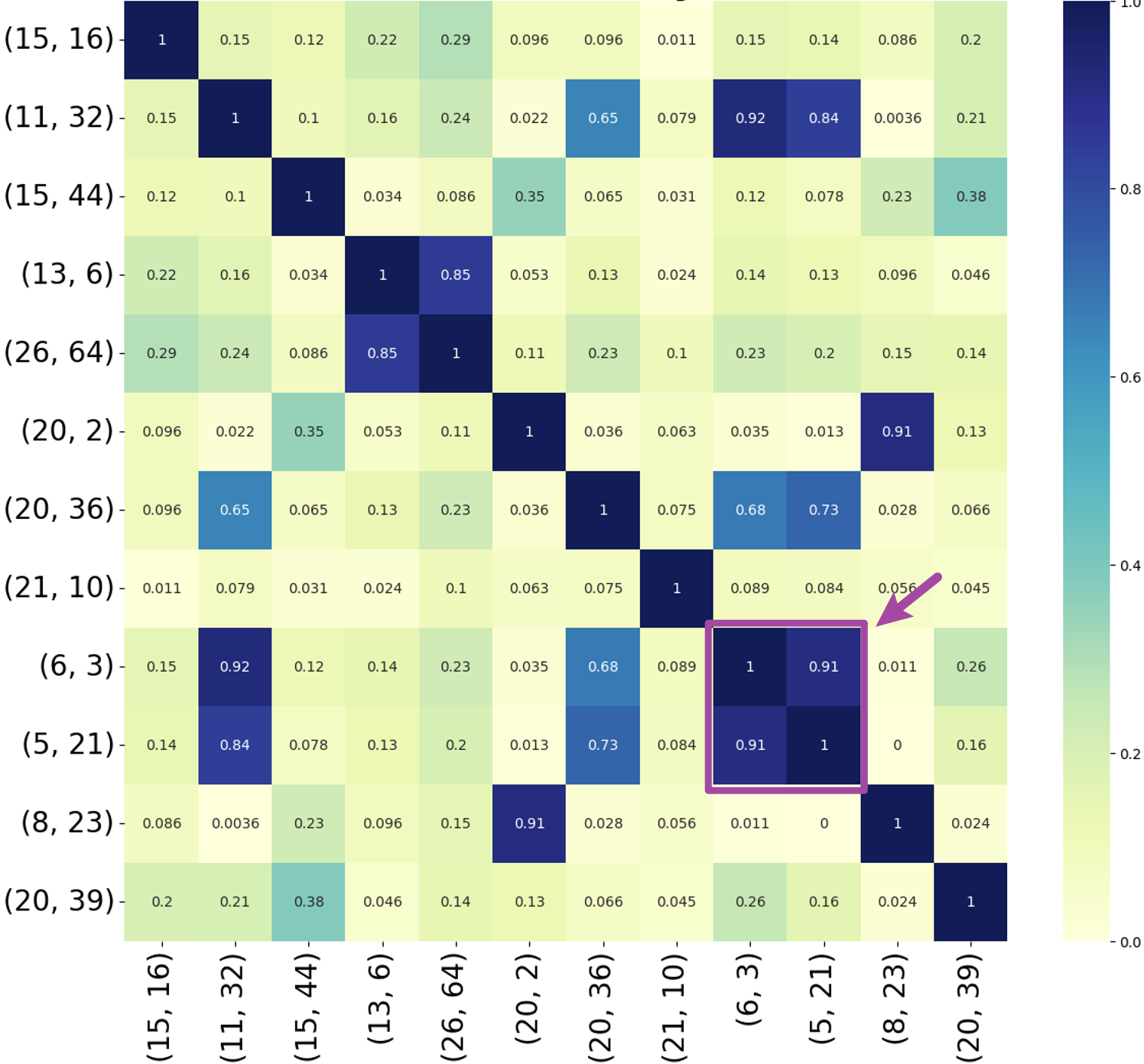}
    \caption{Here ($x,y$) refers to the $y$-th expert in $x$-th layer.  By selecting any two experts from the MoE, we can calculate the probability of their co-activation.  It can be observed that Expert 21 from the layer 5 and Expert 3 from the layer6 frequently activate simultaneously, forming an expert collaboration pattern.}
    \label{fig:introduction_2}
\end{figure}

Despite the widespread success of MoE LLMs, several key challenges remain underexplored. One of the main challenges is understanding the collaborative mechanisms among the experts within the network. While MoE LLMs generate final outputs by combining the predictions of multiple experts,  how these experts cooperate to produce the outputs is still not well understood. 
Figure \ref{fig:expert_collaboration} conceptualizes the notion of cross-layer expert collaboration - coordinated groups of experts across distinct layers that exhibit synchronized activation to implement specific functional modules. This phenomenon is empirically validated in operational MoE networks. Figure \ref{fig:introduction_2}  illustrates a representative case of strong co-activation patterns between Expert 21 in Layer 5 and Expert 3 in Layer 6.
Comprehending these collaboration patterns is essential, as it directly influences knowledge sharing, model interpretability, performance, and optimization. Another key challenge lies in the high model complexity of MoE LLMs, which presents significant challenges in terms of deployment, limiting their scalability for large-scale applications \cite{lu_not_2024,he_demystifying_2024}. 


Therefore, this study aims to investigate and reveal the collaboration patterns between experts in MoE LLMs, and utilize these patterns to enhance model efficiency and performance. The core questions we address include: (1)  Are there consistent collaboration patterns among experts, and what do they reveal about the tasks implicitly learned in MOE LLMs? (2)  Can these collaboration patterns be leveraged to compress MoE LLMs?

\begin{figure*}[ht]
    \centering
    \includegraphics[width=\textwidth]{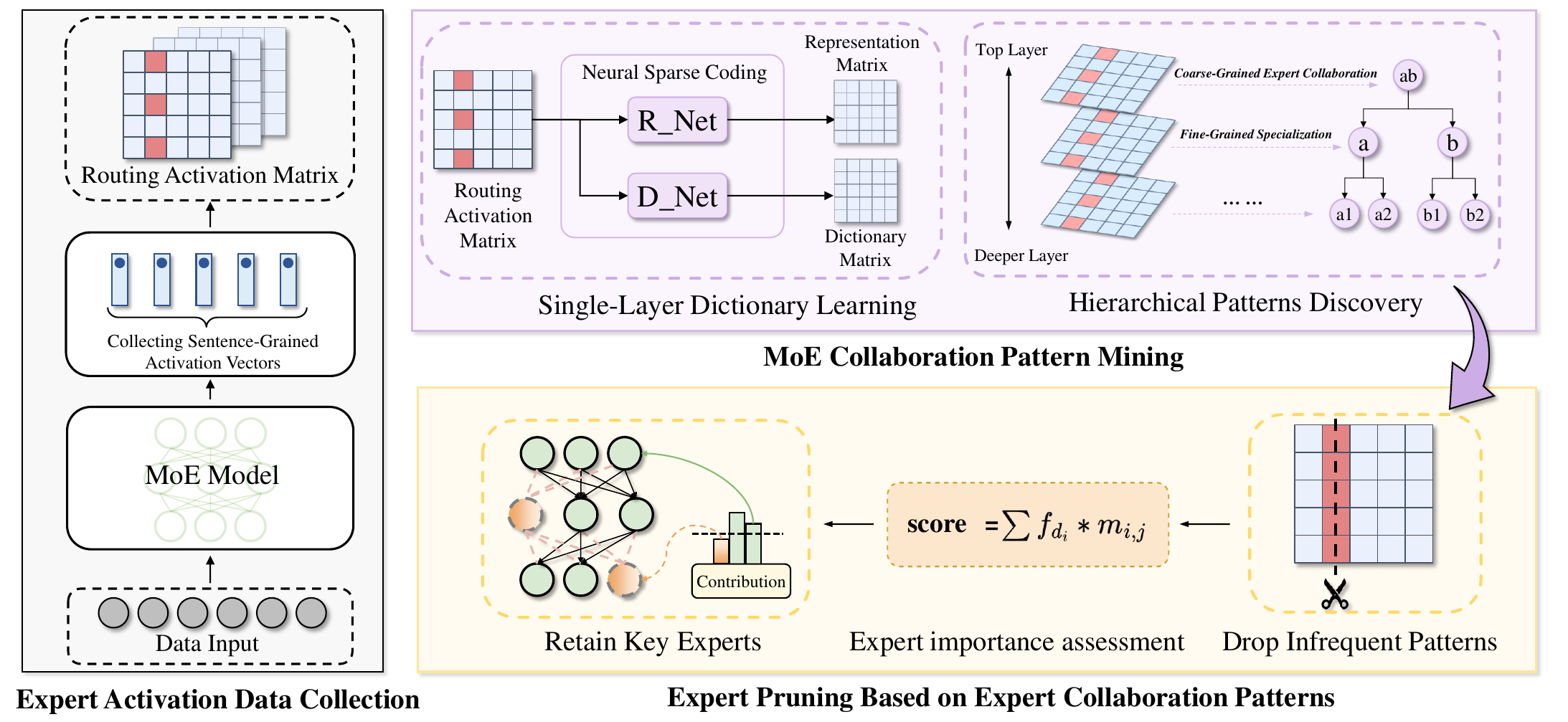}
    \caption{Overview of Our Study's Pipeline.}
    \label{pipeline}
\end{figure*}

To address the two key questions, we begin by extracting the expert activation matrix, which serves as the foundation for further analysis. For the first question, we apply a novel hierarchical sparse dictionary learning (HSDL) approach to uncover collaboration structures within the expert activation data. Building on these insights, we then investigate expert pruning through the Contribution-Aware Expert Pruning (CAEP) algorithm, which identifies and removes low-contribution experts. This process reduces model redundancy, alleviating storage pressure while preserving or even enhancing performance. The entire pipeline, as outlined in Figure \ref{pipeline}, comprises three key components: (1) Expert Activation Data Collection, (2) MoE Collaboration Pattern Mining, and (3) Expert Pruning Based on Expert Collaboration Pattern.


In our experimental evaluation, we tested several representative MoE architectures, including the DeepSeek model, on the MMLU-pro dataset, which contains 2,812 samples across five chosen domains: mathematics, computer science, physics, law, and psychology. Our analysis of the learned dictionaries revealed domain-specific expert collaboration patterns with distinct semantic significance. Building on these insights, we conducted pruning experiments using the CAEP method, which demonstrated that pruning experts based on these patterns effectively reduces the number of experts while maintaining or even improving performance. Our method outperforms baselines with an average improvement of 2.5\%, and in the best case, pruning 50\% of experts results in only a 5.7\% performance drop for specific tasks.


\setlength{\belowcaptionskip}{-0.3cm} 

Our contribution can be summarized as follows:
\begin{itemize}
    \item We explore and uncover the latent collaboration patterns among experts in MoE LLMs. We propose hierarchical sparse dictionary learning (HSDL) and reveal how experts interact and cooperate, which provides new insights into the collaborative mechanisms that drive the performance of MoE LLMs.

    \item We propose the Contribution-Aware Expert Pruning (CAEP) algorithm, which optimizes model efficiency by pruning low-contribution experts without sacrificing performance. Our experiments show that CAEP maintains competitive performance while significantly reducing the number of experts, effectively balancing pruning and performance retention.

\end{itemize}

\section{Literature Review}

\subsection{Analysis of Routing in MoE Networks}

The analysis of router behavior in MoE networks focuses on understanding how the model selects experts based on input features, which is key for optimizing performance. For instance, Lo et al. found that routers typically select experts with larger output norms \cite{lo_closer_2024}, while other studies suggest that router choices are more related to token IDs than to expert fields \cite{jiang_mixtral_2024, xue_openmoe_2024, dai_deepseekmoe_2024}. While these approaches offer valuable insights, they often treat experts as independent entities, overlooking the collaboration patterns between them.


\subsection{Expert Pruning in MoE}

Expert pruning reduces storage consumption in MoE networks by removing less impactful experts. Current strategies include: (1) discarding experts with low activation frequencies based on router decisions \cite{muzio_seer-moe_2024}, (2) identifying experts with minimal output influence using $|x - f(x)|$ differences \cite{lu_not_2024, he_demystifying_2024}, and (3) merging experts by calculating weight similarities \cite{li_merge_2024, zhang_diversifying_2024}. However, these methods often treat experts independently or focus on merging similar groups, without exploring diverse expert combinations with distinct roles.



\subsection{Sparse Dictionary Learning}
Sparse dictionary learning is a well-established method in representation learning and dimensionality reduction \cite{yang_image_2010, wright_robust_2009}. It constructs a dictionary of features that enables sparse representation of data, facilitating efficient encoding of high-dimensional information \cite{tang_explainable_2023, chen_pathlet_2013}. This approach has proven effective in various applications, such as image processing and signal recovery, where it helps capture essential features while reducing noise \cite{hou_detecting_2021, hou_visual_2020}. Recently, companies like OpenAI, Google, and Anthropic have applied sparse dictionary learning to understand large language models' mechanisms \cite{rajamanoharan_improving_2024, gao_scaling_nodate}. Despite its success in other areas, sparse dictionary learning has been underutilized in explanatory research on MoE networks.

\section{Extraction of Expert Activation Matrix} 
In MoE LLMs, the activation weights of the experts reflect the intensity of their responses to the input data, thereby elucidating the collaborative patterns among them. Furthermore, these activation data provide a foundational basis for optimizing pruning strategies, which in turn contribute to enhanced computational and storage efficiency. Consequently, the extraction and analysis of activation weights are critical steps in the effective exploration of collaboration patterns and the implementation of pruning techniques.

Given an MoE LLM with \( m \) layers and \( n \) experts, and an input dataset \( S \) containing \( N_s \) samples, we extract the expert activation data to construct a two-dimensional activation tensor \( V \in \mathbb{R}^{N_s \times (m \times n)} \), where each element \( v_{i,j,k} \) represents the activation weight of the \( k \)-th expert in the \( j \)-th layer for the \( i \)-th sample. This activation weight quantifies the intensity of the expert's response to the input sample, with values constrained within the range \([0, 1]\).

To aggregate the activation data of each sample into a sentence-level representation, we sum the activation values of all tokens within a sample, thereby obtaining the sentence-level activation value for each layer. Let \( \alpha(i)_{t,j,k} \) denote the routing allocation of the \( t \)-th token in sample \( S_i \) to the \( k \)-th expert in the \( j \)-th layer. The sentence-level activation value is then computed as:
\begin{equation}
v_{i,j,k} = \sum_{t=1}^{T} \alpha(i)_{t,j,k}.
\end{equation}
where \( T \) represents the sequence length. Finally, by transposing and accumulating these activation data, we construct the expert activation matrix \( X \), which serves as the input to the subsequent analysis of collaboration patterns among experts.

\section{MoE Collaboration Pattern Mining}

In this section, we propose a novel \textbf{Hierarchical Sparse Dictionary Learning (HSDL)} approach to uncover collaboration patterns among experts in MoE LLMs through hierarchical decomposition. Furthermore, We evaluate its effectiveness on the MMLU-pro dataset, validating the method by comparing it to exhaustive search techniques and exploring domain-specific expert interactions, demonstrating its versatility and efficiency in capturing complex MoE dynamics.

\subsection{Problem Definition}

The objective of this task is to extract the collaboration patterns among experts in MoE LLMs. Given a dataset $S = \{s_1, s_2, \dots, s_{N_s}\}$ comprising $N_s$ samples, we construct an expert activation matrix $X \in \mathbb{R}^{N_e \times N_s}$, where $N_e$ denotes the total number of experts. By employing sparse dictionary learning techniques to decompose $X$, we obtain a dictionary matrix $D \in \mathbb{R}^{N_e \times N_p}$ and a sparse coding matrix $R \in \mathbb{R}^{N_p \times N_s}$, with $N_p$ representing the predefined dictionary capacity. Our goal is to decompose the expert activation matrix \( \mathbf{X} \) into a dictionary matrix \( \mathbf{D} \) and a sparse coding matrix \( \mathbf{R} \), which can be expressed as follows:
\begin{equation}
\mathbf{X} \approx \mathbf{D} \cdot \mathbf{R}.
\end{equation}

Here, the dictionary matrix \( D \) encodes the collaboration patterns among experts, while the sparse coding matrix \( R \) determines how these patterns combine to reconstruct \( X \). 

\begin{figure}[ht]
    \centering
    \includegraphics[width=\columnwidth]{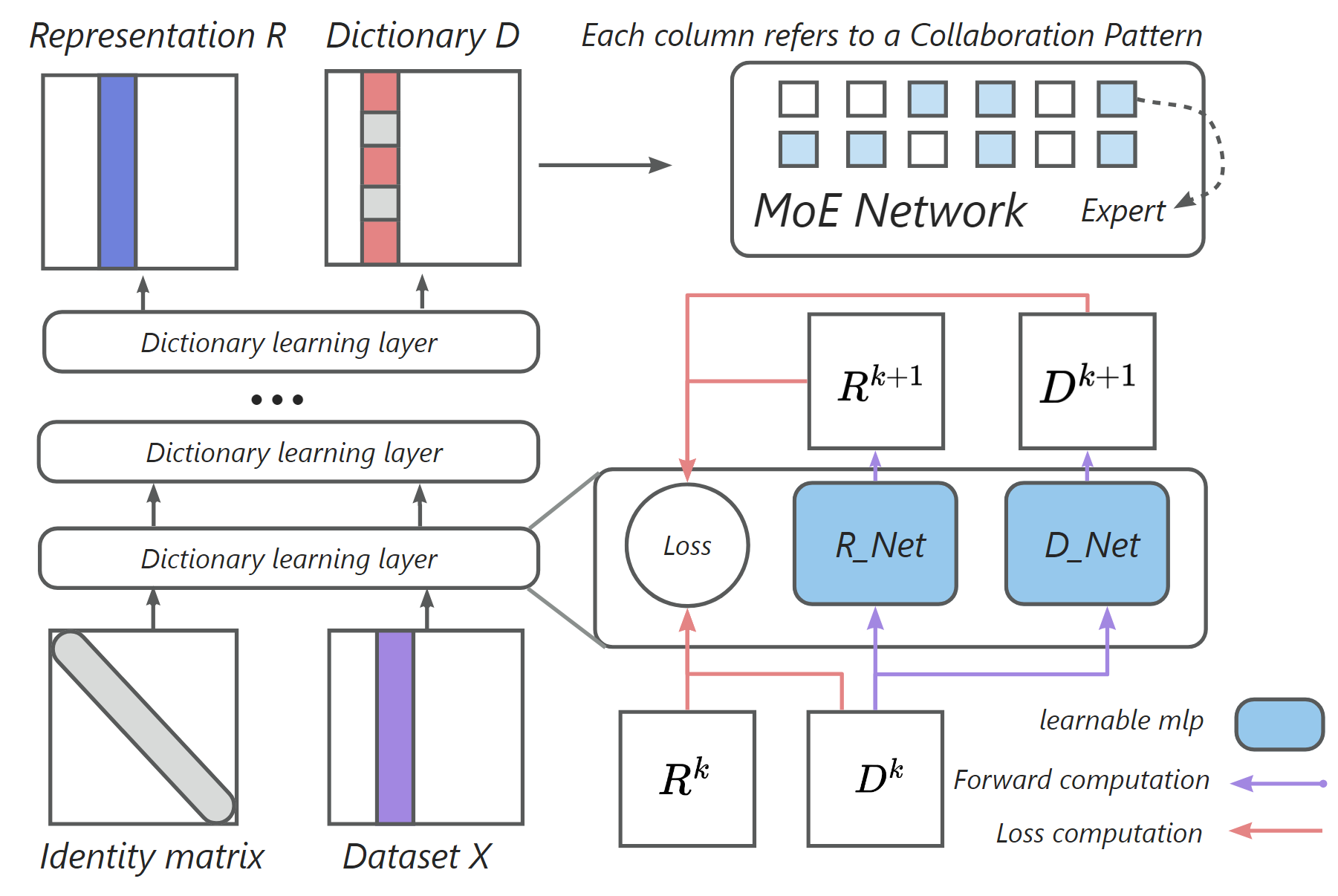}
    \caption{Hierarchical Sparse Dictionary Learning.}
    \label{hierarchical_sparse_dictionary}
\end{figure}

\subsection{Hierarchical Sparse Dictionary Learning for Expert Collaboration Patterns Mining}

Sparse dictionary learning is an effective unsupervised method for uncovering latent structures in data through sparse representations. By modeling data as a linear combination of dictionary atoms, it reveals expert collaboration patterns in MoE LLMs. However, a single-layer approach fails to capture complex patterns across varying granularities. To address this, we propose the HSDL approach, which recursively decomposes the dictionary matrix, capturing collaboration patterns from coarse to fine granularity, thus revealing multi-layered expert interactions.

We extend the original single-layer structure decomposition into a hierarchical structure by recursively decomposing the dictionary matrix at each layer \( k \) into finer subpatterns represented by \( D_{k+1} \), formulated as:
\begin{equation}
D_k \approx D_{k+1} \cdot R_{k+1}.
\end{equation}

Figure \ref{hierarchical_sparse_dictionary} illustrates the hierarchical structure of Sparse Dictionary Learning, showing how the multi-layered expert collaboration is modeled across different layers. 

Furthermore, we introduce three key constraints to optimize the multi-layer dictionary learning process:

(1) \textbf{Sparsity Constraint}: This ensures that the sparse coding matrix \( R_k \) at each layer remains sparse, preventing certain dictionary elements from dominating. Specifically, \( R_{k,i,:} \) denotes the sparse coding of the \( i \)-th data point at layer \( k \). This constraint is defined as:
\begin{equation}
L_{\text{sparse}} = \| R_{k,i,:} \|_\infty.
\end{equation}

(2) \textbf{Inter-Layer Consistency Constraint}: This controls the influence of dictionary learning across layers. The matrix \( R_{k,j} \) represents the contribution of the \( j \)-th dictionary atom at layer \( k \). The formula is:
\begin{equation}
L_{\text{hier}} = \sum_j \| R_{k+1,j} \|_1 \cdot \| R_{k,j} \|_1 / N.
\end{equation}

(3) \textbf{Reconstruction Error Term}: This ensures that the relationships between dictionaries at successive layers are consistently learned. The reconstruction error is defined as:
\begin{equation}
L_{\text{rec}} = \sum_j \| D_{k,j} - (D_{k+1} R_{k+1})_j \|_1 \cdot \| R_{k,j} \|_1 / N.
\end{equation}

These three constraints collectively guide the optimization of both the hierarchical dictionary and sparse coding matrices. The overall loss function is formulated as:
\begin{equation}
L_{\text{total}} = L_{\text{sparse}} + \lambda_1 L_{\text{hier}} + \lambda_2 L_{\text{rec}},
\end{equation}
where \(\lambda_1\) and \(\lambda_2\) are hyperparameters that control the respective losses. By minimizing this loss function, we optimize both the dictionary matrix \( D_k \) and the sparse coding matrix \( R_k \) at each layer, effectively capturing the multi-level structure of expert collaboration.

\subsection{Experimental Analysis of Expert Collaboration Patterns}

In this subsection, we aim to explore how the collaboration patterns among experts in MoE-based LLMs reflect the tasks implicitly learned by the model, thereby contributing to a deeper understanding of its functioning. We present a detailed analysis of the expert collaboration patterns identified through our hierarchical sparse dictionary learning method. To investigate these patterns and their semantic implications, we conduct a series of experiments using the MMLU-pro dataset.

\subsubsection{Experimental Setup}  
We use the phi-moe model and apply our HSDL method to 2,812 samples from the MMLU-pro dataset, covering five domains: mathematics, computer science, physics, law, and psychology.

\subsubsection{Prompt Interpretation using Expert Collaboration Pattern }
To explore how expert collaboration patterns in MoE LLMs reflect the model’s understanding of tasks, we conduct a detailed analysis using the hierarchical dictionary learning method. Specifically, we aim to understand how different experts collaborate to handle specific aspects of a problem.

\begin{figure}[ht]
    \centering
    \includegraphics[width=\columnwidth]{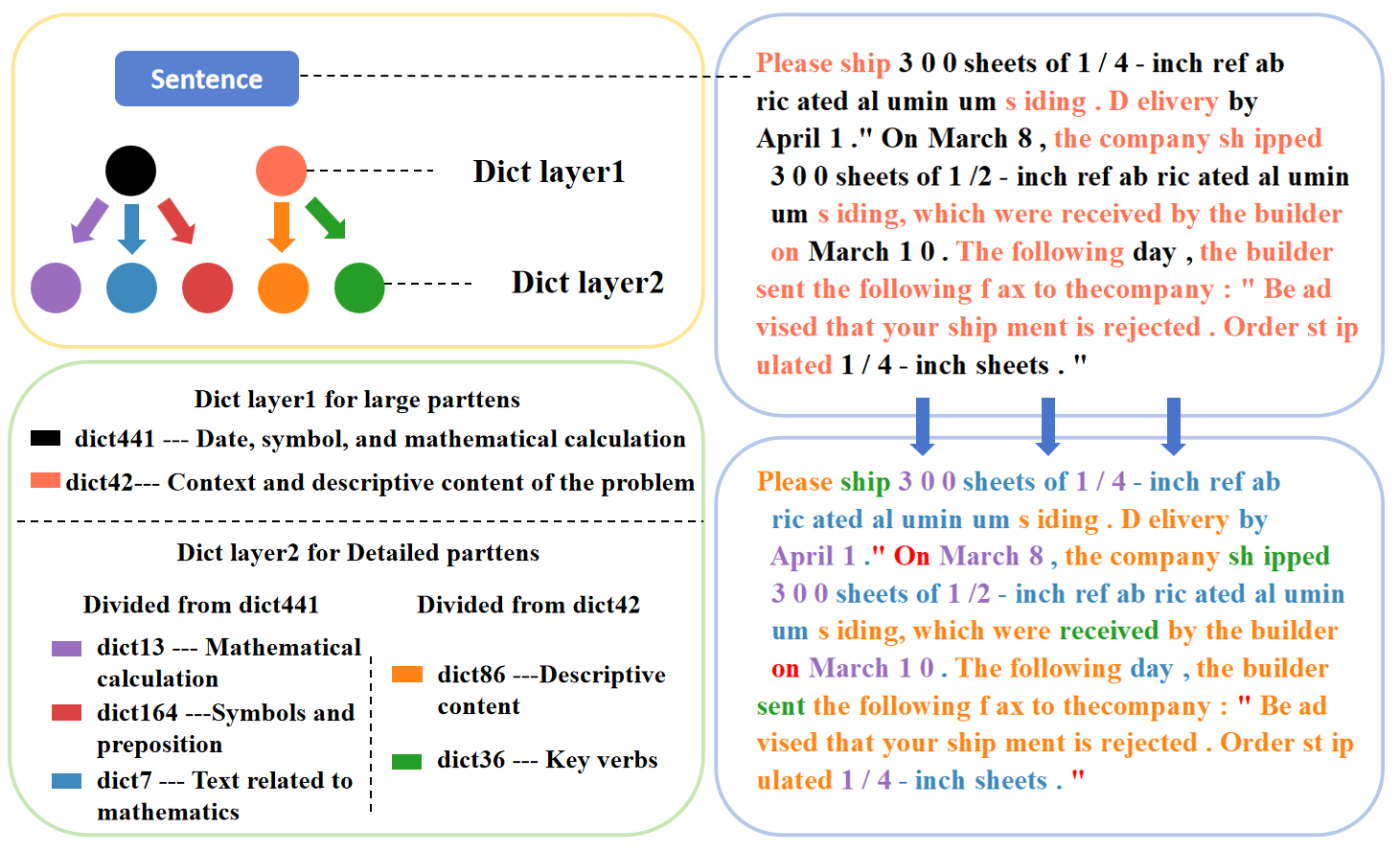}
    \caption{Hierarchical Semantic Annotation of Dictionary Elements on MMLU.}
    \label{Semantic_Annotation}
\end{figure}

To achieve this, we designed a semantic annotation scheme for input sentences to interpret the semantics of expert collaboration patterns derived from HSDL. We color words processed by the same dictionary atoms (i.e., expert collaboration patterns) with the same color. This color-coding scheme facilitates the observation of both the performance and interrelationships of the expert collaboration patterns. We analyze the input samples using the dictionary atoms obtained through HSDL, with one such analysis shown in Figure \ref{Semantic_Annotation}.

\textbf{Results and Discussion.} We find that the hierarchical semantic annotation of expert collaboration patterns reveals how MoE LLMs understand and process different tasks within a problem. As shown in Figure \ref{Semantic_Annotation}, in the upper left corner, we can observe that:  \textbf{Expert collaboration patterns in higher-layer and lower-layer dictionaries demonstrate a hierarchical  semantic relationship, which becomes increasingly fine-grained as layer increases.
} The lower left corner of the figure displays this from a semantic perspective, where the top layer captures broad categories such as "Date, symbol, and mathematical calculation," while deeper layers break these down into more detailed components like "Mathematical calculation" or "Key verbs."

These findings provide a direct answer to our central question on expert collaboration patterns in MoE LLMs. The hierarchical decomposition offers a more detailed understanding of the model's internal processes, shedding light on how tasks are learned and executed.  This approach could evolve into a tool for visualizing MoE LLMs behavior, enhancing interpretability and supporting optimization for domain-specific applications.
\subsubsection{Comparison with Exhaustive Search Results}

To investigate whether the top dictionary elements correspond to the most frequent expert combinations, we compared the dictionary's expert collaboration patterns with those from an exhaustive search method. Due to the high computational cost of considering larger combinations, we limited the analysis to pairs and triplets.

To quantify the coverage of the most frequent expert combinations in our dictionary, we define \( N_{\text{top}} \) as the number of dictionary items in the top \( k\% \) of the traversal pattern, and \( N_{\text{total}} \) as the total number of dictionary items. The coverage is then calculated using the following formula:
\begin{equation}
\text{Top } k\% \text{ Coverage} = \frac{N_{\text{top}}}{N_{\text{total}}}.
\end{equation}

\begin{figure}[ht]
    \centering
    \includegraphics[width=\columnwidth]{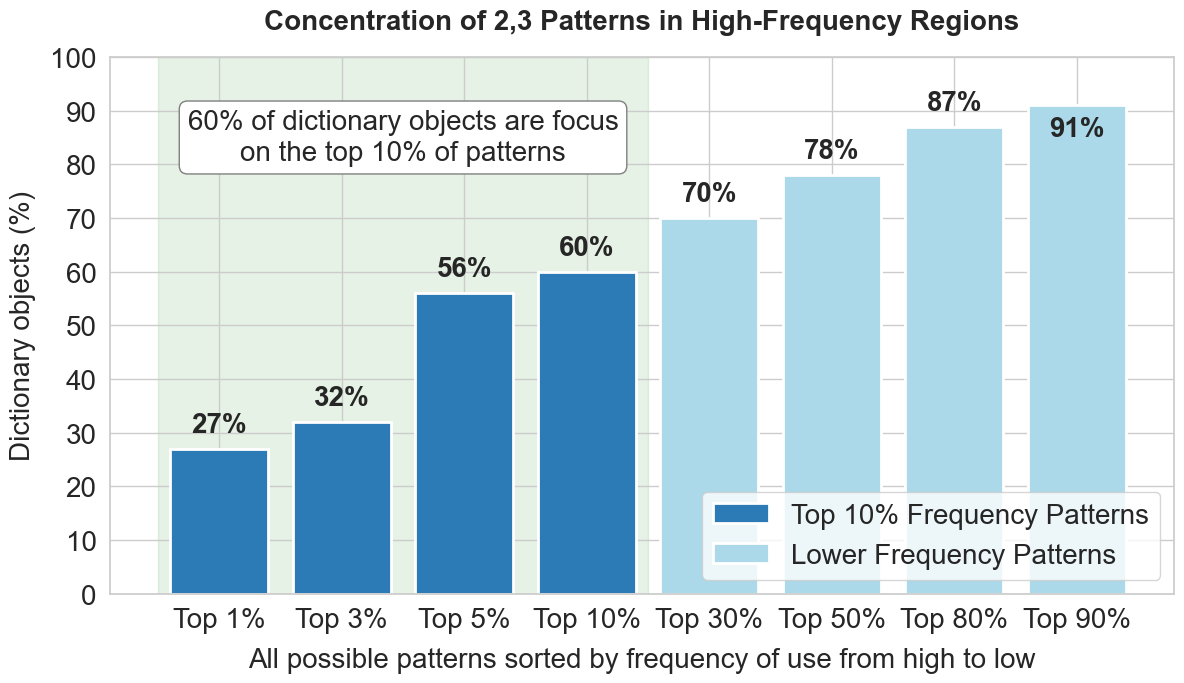}
    \caption{Comparison of overlap with the results of the exhaustive method.}
    \label{comparision}
\end{figure}

\textbf{Results and Discussion.} As shown in Figure \ref{comparision}, \textbf{the collaboration patterns identified by our method predominantly align with the most frequent expert combinations found during the exhaustive search.} Specifically, 60\% of the patterns identified by our method correspond to the top 10\% of the most frequent expert combinations, indicating that our method efficiently identifies the most prevalent collaboration patterns.

While our method focuses on the most frequent expert combinations, it also captures some low-frequency patterns. These less frequent combinations, though less common, are critical for capturing the diversity of expert interactions, which enhances the model’s ability to tackle a wider range of tasks. \textbf{This highlights the importance of considering both high- and low-frequency expert combinations in shaping the performance and versatility of MoE LLMs.}

\subsubsection{Domain-Specific Expert Collaboration Patterns}
In this experiment, our goal is to explore how expert collaboration patterns vary across different domains and to understand the domain-specific nature of expert interactions within MoE LLMs. Specifically, we aim to examine the activation frequencies of experts for inputs from various fields, including mathematics, computer science, physics, law, and psychology, to uncover potential domain-related patterns.

we analyzed the frequency distribution of activated experts during the model processing for inputs from different domains and calculated the cosine similarity between the distributions of each domain, resulting in a confusion matrix.

\begin{figure}[ht]
    \centering
    \includegraphics[width=\columnwidth]{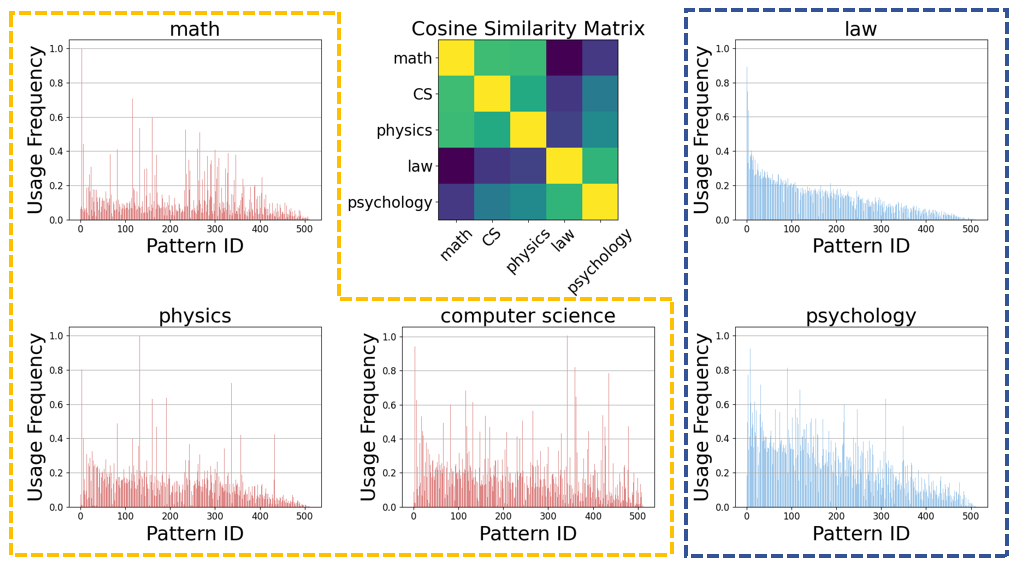}
    \caption{The distribution of expert selection frequencies during inputs from different fields.}
    \label{distribution}
\end{figure}

\textbf{Results and Discussion.} Figure \ref{distribution} shows the expert selection frequency distribution across domains. We can observe that for inputs from different fields, the distribution of expert activation frequencies in the MoE LLM varies. For semantically similar domains, such as mathematics, physics, and computer science indicated by the orange dashed box, their distributions are closer to each other. In contrast, the distributions of expert activation frequencies are more different for domains with greater semantic differences, such as mathematics and law. \textbf{This suggests that expert collaboration is more specialized within specific domains, reflecting domain-specific interactions in MoE LLMs.}

These findings indicate that \textbf{experts in MoE LLMs exhibit domain preferences, adjusting expert selection based on the input domain's characteristics to optimize performance for domain-specific tasks.} Understanding these patterns can enhance the model's efficiency and its ability to handle specialized tasks.



\section{Expert Pruning Based on Expert Collaboration Patterns}
In this section, we present the CAEP method, which utilizes expert collaboration patterns to reduce the number of experts in an MoE LLM while preserving performance. We first introduce the pruning algorithm and then demonstrate its effectiveness through two types of experiments: (1) General Tasks Evaluation, where we compare CAEP with baseline methods on diverse tasks, and (2) Domain-Specific Evaluation, where we assess its ability to retain domain-relevant capabilities after pruning.



\subsection{Pruning algorithm}
We propose the \textbf{Contribution-Aware Expert Pruning (CAEP)} algorithm. The algorithm aims to produce a mask vector that incorporates our retention strategy, given a specific pruning ratio \( k \). In this mask vector, experts corresponding to positions with a value of 1 are retained, while those with a value of 0 are discarded.
This pruning process is achieved by progressively discarding less significant dictionary atoms, guided by the contribution scores derived from \( R \). The CAEP algorithm proceeds as follows (Algorithm~\ref{alg:caep}):
\begin{itemize}
    \item \textbf{Calculation and Ranking}: Calculate the contribution scores for each expert by the sparse representation matrix \( R \) and the dictionary matrix \( D \), obtaining the total contribution and sorting it in descending order.
    \item \textbf{Initial Threshold Mask}: Determine the score based on the predefined threshold ratio and generate the initial binary mask, marking the experts whose contribution scores are above.
    \item \textbf{Iterative Pruning}: Before reaching the target pruning ratio, repeatedly identify the least used patterns and remove them from the dictionary and the sparse representation while updating the contribution scores and the mask, until only the desired ratio of experts remains.
\end{itemize}


\begin{algorithm}[htbp]  
\small  
\caption{Expert Pruning Strategy}  
\label{alg:caep}  
\begin{algorithmic}[1]  
\Require   
    Dictionary matrix $\mathbf{D} \in \mathbb{R}^{N_e \times N_p}$ \\  
    Sparse representation matrix $\mathbf{R} \in \mathbb{R}^{N_p \times N_s}$ \\  
    Threshold ratio $k_1 \in (0,1)$ \\  
    Target pruning ratio $k_2 \in (0,1)$  
\Ensure Pruned expert mask $\mathbf{m} \in \{0,1\}^{N_e}$

\State $\mathbf{R}_{\text{sum}} \gets \sum_{j=1}^{N_s} \mathbf{R}_{:,j}$ \Comment{Sum over samples}  
\State $\mathbf{D}_{\text{sum}} \gets \mathbf{D} \cdot \mathbf{R}_{\text{sum}}^{\top}$ \Comment{Weighted by pattern frequency}  
\State $\mathbf{e} \gets \sum_{i=1}^{N_p} \mathbf{D}_{\text{sum},i}$ \Comment{Aggregate expert contributions}

\State Sort $\mathbf{e}$ in descending order: $\mathbf{e}_{\text{sorted}}$  
\State $f \gets \mathbf{e}_{\text{sorted}}[\lceil k_1 \cdot N_e \rceil]$ \Comment{Threshold at $k_1$-quantile}  
\State $\mathbf{m} \gets \mathbf{1}_{\mathbf{e} \geq f}$ \Comment{Initial binary mask}  
 
\While{$\|\mathbf{m}\|_0 > (1 - k_2) \cdot N_e$}  
    \State $i^* \gets \arg\min_{i} \mathbf{R}_{\text{sum}}(i)$ \Comment{Find least used pattern}  
    \State Remove column $i^*$ from $\mathbf{D}$ and row $i^*$ from $\mathbf{R}$  
    \State Recompute $\mathbf{R}_{\text{sum}}$, $\mathbf{D}_{\text{sum}}$, $\mathbf{e}$  
    \State Update $\mathbf{m} \gets \mathbf{1}_{\mathbf{e} \geq f}$ \Comment{Adapt mask}  
\EndWhile  

\Return $\mathbf{m}$  
\end{algorithmic}  
\end{algorithm} 

\subsection{Experiments on General and Domain-Specific Tasks}
We conduct a series of experiments to evaluate the effectiveness of our proposed pruning method, CAEP. We perform experiments on both general tasks and domain-specific tasks. The goal is to assess how well the pruned model retains its capabilities across a variety of tasks, while optimizing performance retention in specific domains. 
The dataset and specific configurations used in this part of the experiment can be found in Appendix B.

\subsubsection{Experiments on General Tasks}
The goal of this experiment is to evaluate how well the pruned model retains its performance across a broad set of general tasks. We compare CAEP with baseline pruning methods to analyze the trade-off between reducing the number of experts and maintaining task performance.

\textbf{Comparison with Other Expert Pruning Baselines.}
We compare CAEP to two baseline pruning strategies: (1) Routing Score-Based Pruning \cite{muzio_seer-moe_2024}: Retains experts with higher averaged routing scores. (2) Behavior-based Pruning \cite{zhang_diversifying_2024}: Remove experts with minimal impact on the output. 


\begin{table*}[ht] 
\centering
\caption{Performance evaluation of different expert pruning methods with 25\% experts dropped.}
\label{tab:Evaluate}
\scriptsize 
\resizebox{\linewidth}{!}{%
\begin{tabular}{llccccccc}
\toprule
\textbf{Model} & \textbf{Method} & \textbf{AVG↑} & \textbf{OBQA↑} & \textbf{ARC-C↑} & \textbf{HellaSwag↑} & \textbf{WinoGrande↑} & \textbf{RTE↑} & \textbf{PIQA↑} \\
\midrule
\multirow{4}{*}{DeepSeek} & Random & 0.500 & 0.363 & 0.564 & 0.485 & 0.568 & 0.641 & 0.381 \\
 & SEER-MoE & \underline{0.5872} & 0.420 & \underline{0.672} & \underline{0.665} & 0.617 & \underline{0.755} & \underline{0.394} \\
 & GEM & 0.5870 & \underline{0.422} & 0.67 & 0.658 & \textbf{0.649} & 0.739 & 0.384 \\
 & \textbf{CAEP (Ours)} & \textbf{0.612} & \textbf{0.473} & \textbf{0.693} & \textbf{0.691} & \underline{0.635} & \textbf{0.757} & \textbf{0.424} \\
\bottomrule
\end{tabular}%
}
\end{table*}

\begin{figure}[ht]
    \centering
    \includegraphics[width=\columnwidth]{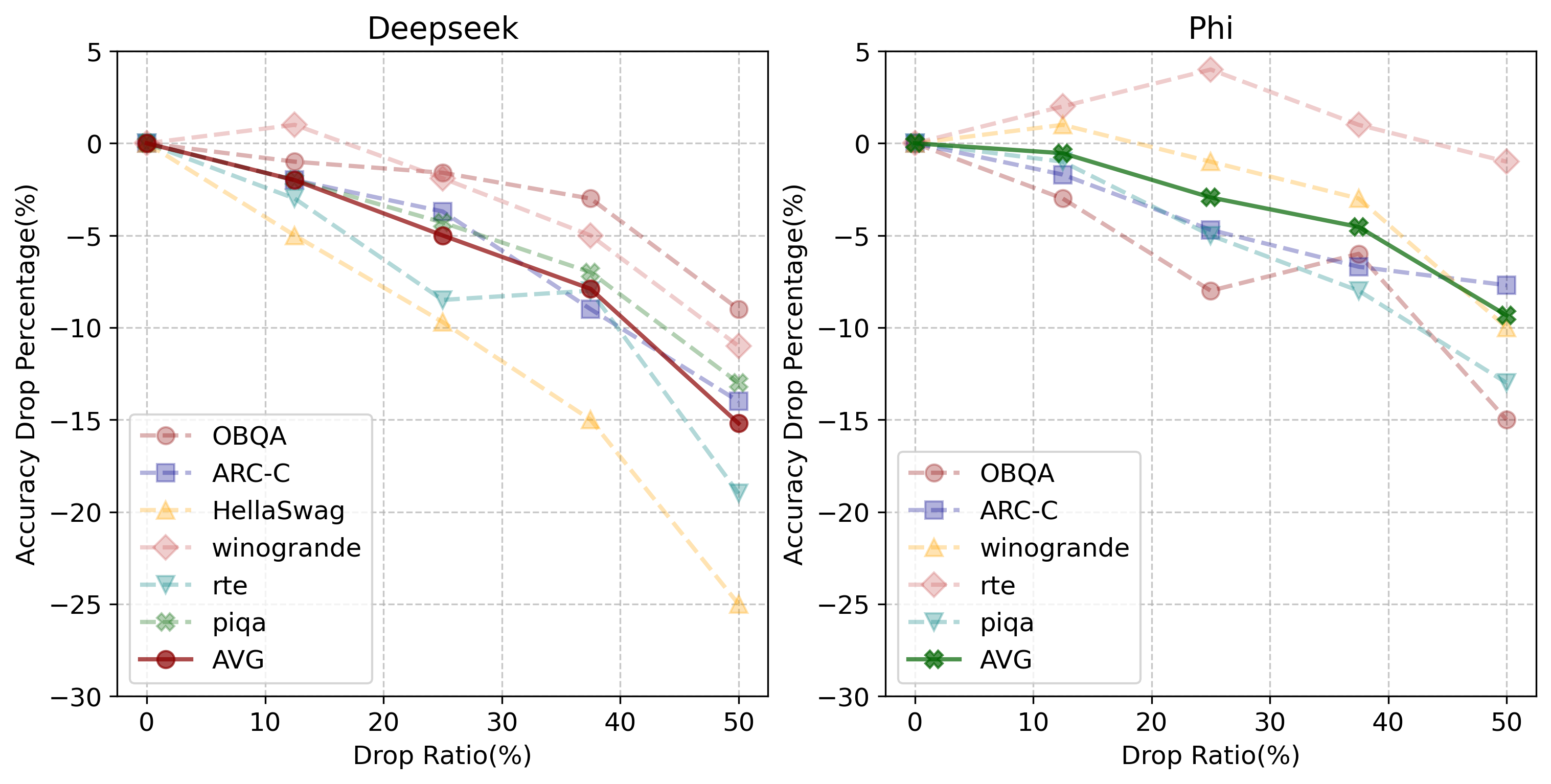}
    \caption{Performance of CAEP on benchmark tasks with varying expert pruning drop ratios.}
    \label{fig:benchmark_comparison}
\end{figure}

\textbf{Results and Discussion.} Figure \ref{fig:benchmark_comparison} and Table \ref{tab:Evaluate} show that CAEP-pruned models maintain competitive performance, outperforming random and baseline methods with an average score of 0.612. Notably, CAEP retains higher performance after pruning 25\% of the experts, especially on tasks like OBQA and RTE. This is further supported by Figure \ref{fig:benchmark_comparison}, where CAEP shows a lower accuracy drop across multiple tasks, including HellaSwag and PIQA, even with a high pruning ratio.

Through the analysis of the experimental results, we found that CAEP effectively retains performance across a broad set of general tasks while significantly reducing the number of experts. This demonstrates that \textbf{CAEP successfully balances pruning and performance retention, optimizing computational efficiency while minimizing performance loss.} 

\renewcommand{\arraystretch}{1.2} 

\useunder{\uline}{\ul}{}

\subsubsection{Experiments on Domain-Specific Tasks}
In this experiment, we focus on investigating how expert collaboration patterns differ across various domains and how these differences reflect the domain-specific interactions within MoE LLMs. Our objective is to analyze the activation frequencies of experts for inputs from five fields—mathematics, computer science, physics, law, and psychology—in order to identify domain-dependent patterns in expert selection.For each domain, we prune 50\% of the experts using CAEP on Phi. This setup enables us to assess whether the pruned model retains superior performance in a specific domain at the expense of others.

\textbf{Performance Evaluation Metric.}  
To assess the impact of pruning, we focus primarily on the relative changes in performance. The metric is computed as:
\begin{equation}
\frac{Acc_{\text{pruned}} - Acc_{\text{no-pruned}}}{Acc_{\text{no-pruned}}}.
\end{equation}
A higher value indicates better retention of domain-specific capabilities, with the ideal result being maximized diagonal elements, showing that each pruned model retains domain-specific expertise.

\begin{figure}[ht]
    \centering
    \includegraphics[width=0.8\columnwidth]{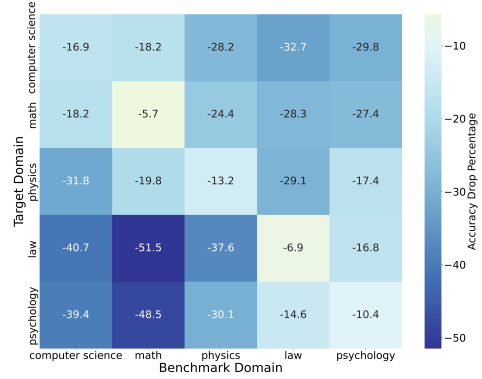}
    \caption{Performance degradation accuracy after pruning for specific domain}
    \label{fig:heatmap}
\end{figure}

\textbf{Results and Discussion.} Figure \ref{fig:heatmap} shows the accuracy degradation after pruning for different domains, presented as a heatmap. The color scale indicates the percentage of accuracy drop, where darker blue shades represent larger losses. From the figure, we observe that pruning for domains like law and psychology leads to the most significant accuracy drops, particularly when the target domain is law. In contrast, pruning for the "physics" or "psychology" domains results in relatively smaller accuracy drops, suggesting a less severe impact on performance.

\textbf{We find that this variation in pruning impact, depending on both the target and benchmark domains, reveals an uneven distribution of domain-specific knowledge across the model.} Some domains rely more heavily on specialized expertise, while others are more flexible in terms of expert collaboration. These findings suggest that \textbf{pruning strategies should account for the varying importance of domain-specific knowledge, allowing for more efficient expert retention and minimizing unnecessary performance degradation in MoE LLMs.}


\section{Conclusion}

This paper addresses a key gap in MoE LLMs, where existing research has largely overlooked the collaboration patterns among experts, both within the same layer and across layers. By applying hierarchical sparse dictionary learning, we uncover dominant expert collaboration patterns and develop a pruning strategy to enhance MoE LLMs' efficiency. Our experiments demonstrate that this approach not only improves accuracy but also significantly boosts model compression and inference efficiency compared to existing methods. This work provides valuable insights into expert interactions and offers a novel way to optimize MoE LLMs for both performance and scalability.

\newpage

\bibliography{moe_reference, references_1}

\newpage

\appendix
\section*{Appendix} 
\section{Limitations and Future Work}  
The entire work operates under the assumption that the allocation result provided by the router is the most optimal. However, this may only reflect one aspect of the model's behavior. By considering both router information and weight data in a more comprehensive way, we could gain a deeper and more complete understanding. Additionally, there has been limited analysis from the perspective of combinatorial learning, which might offer useful insights into the task selection process. Moreover, the labeling of mined patterns has primarily been done manually up until now. In the future, we aim to explore automating this process, such as using large language models to identify and summarize common tokens associated with specific expert collaboration patterns.


\section{Pruning Effect Calculation}
For the DeepSeek-MoE-16B model, considering the significant impact of shared experts on the model, we only prune the normal experts during the pruning operation. Through calculations, we estimate the parameter counts of various parts of DeepSeek-MoE-16B as follows: word embeddings 0.2B, attention mechanism 0.4B, gate and shared experts 0.9B, routing network of MoE 14.7B, and output layer 0.2B. Therefore, for this model, our conclusion is that the total parameters after pruning with a pruning ratio of \( k\% \) can be calculated as:

\begin{equation}
\text{New Total Parameters} = (16.4 - 14.7 \times k\%) \, \text{B}
\end{equation}

\section{Pruning Experiment Setup}
In section 5, following the setup in \cite{he_demystifying_2024}, we implement our pruning method on the MMLU \cite{hendrycks2021measuring} dataset, using 128 samples with an input sequence length of 2,048 tokens. All pruning experiments are conducted on the DeepSeek-MoE-16B model, where only normal experts are pruned, preserving shared experts due to their importance. Model performance is evaluated using the LM-Harness benchmark, which includes a range of tasks: ARC-C \cite{clark2018think}, BoolQ \cite{clark2019boolq}, HellaSwag \cite{zellers2019hellaswag}, MMLU
, OBQA \cite{mihaylov2018suit}, PIQA \cite{bisk2019piqa}, RTE \cite{wang2019glue}, and WinoGrande \cite{ai2:winogrande}. The evaluation is carried out using the EleutherAI LM Harness framework \cite{eval-harness}, and we report normalized zero-shot accuracy for each task.

\end{document}